\documentclass[letterpaper, 10 pt, conference]{ieeeconf}

\IEEEoverridecommandlockouts

\usepackage{color}
\usepackage{multirow}
\usepackage{booktabs}
\usepackage{url}
\usepackage{subfigure}
\usepackage{threeparttable}
\usepackage{cite}
\makeatletter
\let\NAT@parse\undefined
\makeatother
\usepackage{hyperref}[colorlinks,linkcolor=blue]
\usepackage[ruled,linesnumbered]{algorithm2e}
\renewcommand{\baselinestretch}{0.966}
\usepackage[margin=1in]{geometry}
\usepackage{array}
\newcolumntype{L}[1]{>{\raggedright\arraybackslash}p{#1}}
\usepackage{hyperref}[colorlinks,linkcolor=blue]
\usepackage[ruled,linesnumbered]{algorithm2e}
 \renewcommand{\baselinestretch}{0.966}

\usepackage{booktabs}
\usepackage{enumerate}
\usepackage{amssymb}
\usepackage{leftidx}
\usepackage{amsfonts}
\usepackage{amsmath}
\usepackage{bm}
\usepackage{booktabs}
\usepackage{setspace}
\usepackage{makecell}
\usepackage{setspace}
\usepackage{siunitx}
\usepackage{float}
\usepackage{subfigure}
\usepackage[pdftex]{graphicx}
\usepackage{cite}
\usepackage{siunitx}

\graphicspath{{figs/}}
\usepackage{graphicx}
\graphicspath{{figs/}}
\newcommand{\etal}{\textit{et al}.}
\newcommand{\ie}{\textit{i}.\textit{e}.}
\newcommand{\eg}{\textit{e}.\textit{g}.}

\newcommand{\unknown}{$\mathtt{Unknown}$}
\newcommand{\knownfree}{$\mathtt{KnownFree}$}
\newcommand{\occupied}{$\mathtt{Occupied}$}
\newcommand{\inflatedoccupied}{$\mathtt{InflatedOccupied}$}
\DeclareSIUnit{\mps}{m/s} 

\begin{document}

\title{\LARGE \bf

ROG-Map: An Efficient Robocentric Occupancy Grid Map for Large-scene and High-resolution LiDAR-based Motion Planning}

\author{Yunfan~Ren, Yixi Cai, Fangcheng~Zhu, Siqi Liang and Fu~Zhang
 \thanks{Y. Ren, Y. Cai, F. Zhu, and F. Zhang are with the Department of Mechanical Engineering, University of Hong Kong
 \texttt{\{renyf, yixicai, zhufc\}}\texttt{@connect.hku.hk},
 \texttt{\{fuzhang\}}\texttt{@hku.hk},
 S. Liang is with School of Mechanical Engineering and Automation, Harbin Institute of Technology 
 \texttt{sqliang@stu.hit.edu.cn}.
}
}

\maketitle
\pagestyle{empty} 
\thispagestyle{empty} 

\begin{abstract}
Recent advances in LiDAR technology have opened up new possibilities for robotic navigation. Given the widespread use of occupancy grid maps (OGMs) in robotic motion planning, this paper aims to address the challenges of integrating LiDAR with OGMs. To this end, we propose ROG-Map, a uniform grid-based OGM that maintains a local map moving along with the robot to enable efficient map operation and reduce memory costs for large-scene autonomous flight. Moreover, we present a novel incremental obstacle inflation method that significantly reduces the computational cost of inflation. The proposed method outperforms state-of-the-art (SOTA) methods on various public datasets. To demonstrate the effectiveness and efficiency of ROG-Map, we integrate it into a complete quadrotor system and perform autonomous flights against both small obstacles and large-scale scenes. During real-world flight tests with a \SI{0.05}{m} resolution local map and \SI{30}{\meter}$\times$\SI{30}{\meter}$\times$\SI{6}{\meter} local map size, ROG-Map takes only \SI{29.8}{\percent} of frame time on average to update the map at a frame rate of \SI{50}{\hertz} (\ie, \SI{5.96}{ms} in \SI{20}{ms}), including \SI{0.33}{\percent} (i.e., \SI{0.66}{ms}) to perform obstacle inflation, demonstrating outstanding real-world performance. We release ROG-Map as an open-source ROS package\footnote{https://github.com/hku-mars/ROG-Map} to promote the development of LiDAR-based motion planning.

\end{abstract}

\section{Introduction}
\label{sec:intro}

LiDAR-based autonomous drones have seen significant advancements in various applications, such as search and rescue, inspection, and autonomous exploration. Compared to depth cameras with a sensing range of about $3\sim5$\SI{}{m}, LiDAR sensors provide more precise and long-range (typically ranging from tens to hundreds of meters) three-dimensional measurements, extending the perception range for autonomous UAVs. Additionally, precise LiDAR points enable autonomous drones to avoid small obstacles \cite{kong2021avoiding}, making it possible for them to operate in more challenging areas.

Occupancy mapping is an essential component of an autonomous aerial system to navigate in unknown environments. The occupancy grid map (OGM) is one of the most promising map structures for real-time occupancy mapping, as verified in \cite{zhou2020ego,ren2022bubble,zhou2020robust}, with three critical abilities carefully designed for motion planning as follows. Firstly, ray casting is used to distinguish the occupied, free, and unknown regions of the environment. Secondly, an inflation technique is applied to the occupied space to guarantee safety in consideration of the drone's size. Finally, OGM with high enough resolution is aware of small obstacles, enabling navigation and obstacle avoidance in complex environments. 

Recent research in LiDAR-based motion planning has posed new challenges. To fully utilize the capabilities of LiDAR, which provides long-range and accurate measurements, an ideal occupancy grid map (OGM) for LiDAR is expected to maintain high resolution and update over a long distance with high efficiency. However, these requirements result in significantly increased computation load for map updates and obstacle inflation. As a result, traditional octree-based \cite{hornung2013octomap,duberg2020ufomap,funk2021multi} and hashing-based \cite{niessner2013real} OGMs encounter great difficulties in their real-time ability for LiDAR-based motion planning with limited onboard computation resources. An alternative is the uniform grid-based approach \cite{zhou2020ego,ren2023online,zhou2021fuel} which has high computational efficiency. However, the extensive memory consumption of uniform grid maps makes them impractical for large-scale environments.

\begin{figure}[t]
\centering 
\includegraphics[width=0.48\textwidth]{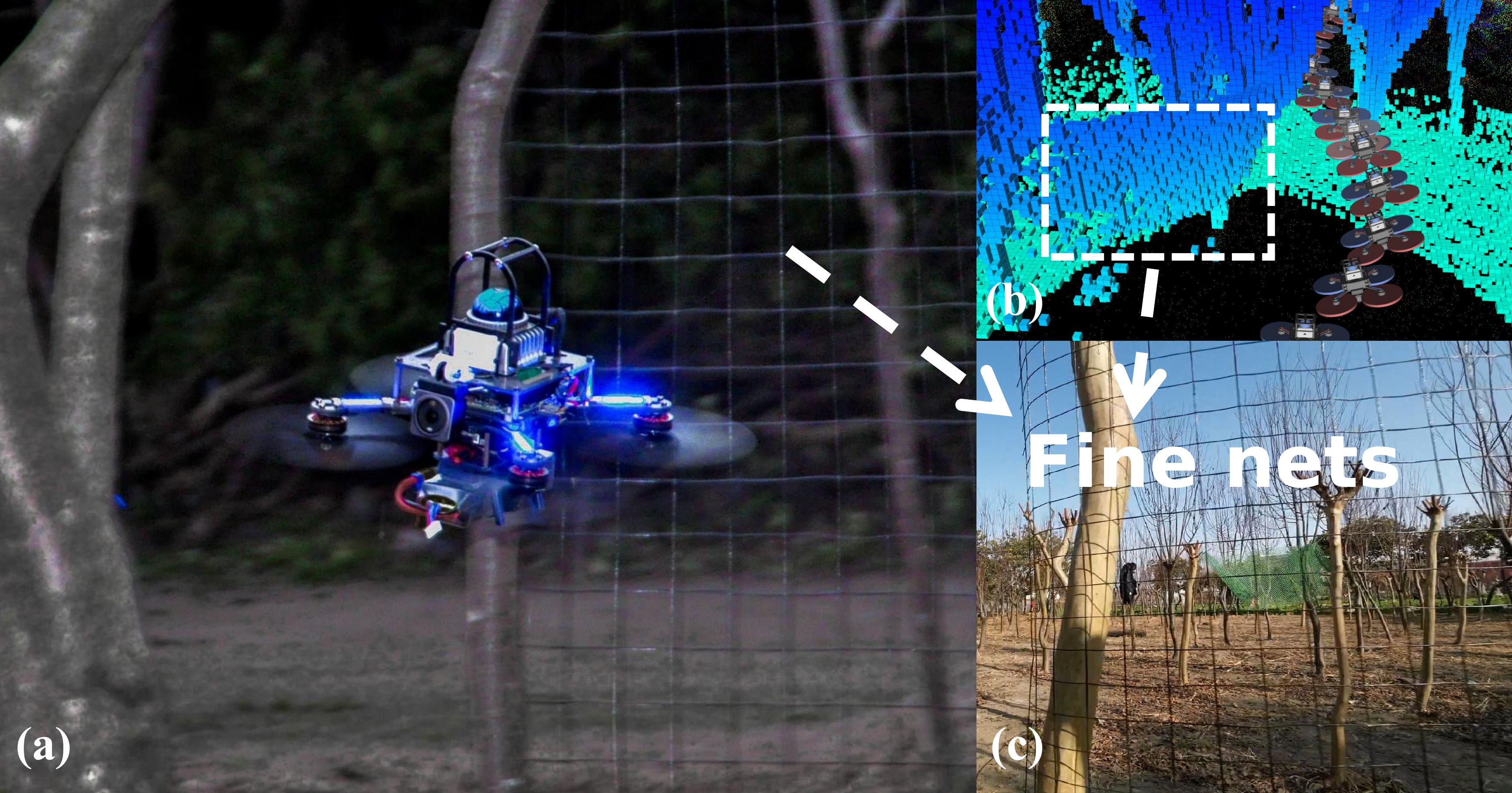}
\vspace{-0.3cm}
 \caption{(a) Fast autonomous navigation against fine metal nets in cluttered environments using Livox Mid360 LiDAR. The FPV camera is only for visualization. (b) The occupied grids in ROG-Map with a resolution of \SI{0.05}{m} during the flight. (c) The fine wire nets with a width of \SI{3}{mm}. The drone is equipped with fully onboard sensing, computation, and control. More details can be found in the attached video at https://youtu.be/eDkwGXCea7w.
}
\label{fig:cover}
\end{figure}

To address the issues in LiDAR-based motion planning, we propose \textbf{ROG-Map}, a uniform grid-based OGM, which is computationally efficient in updating map. Further, by extending \cite{Fankhauser2016GridMapLibrary} to the three-dimensional cases, we propose a zero-copy map sliding strategy that only maintains a local map around the robot, making ROG-Map suitable for large-scene tasks. In addition, we introduce a novel incremental inflation method that significantly reduces the time required for obstacle inflation thereby enhancing overall performance. In summary, the contributions of this paper are as follows:

\begin{itemize}
 \item [1)] 
We propose ROG-Map, a computation-efficient OGM package based on uniform grids. Its zero-copy map sliding strategy enables it to maintain a local map moving with the robot, making it suitable to operate at high resolution and in large-scale environments.

 \item [2)] 
We propose a novel incremental inflation method relying on the change of occupancy state, which achieves a computational complexity of $O(n)$ in the number of changed grids. This method enables faster obstacle inflation with consistent accuracy as existing methods. 
 
 \item [3)]
We compare the ROG-Map against SOTA baselines on public datasets, demonstrating its computation and memory efficiency superiority. Additionally, ROG-Map is integrated into a LiDAR-based quadrotor to conduct extensive real-world tests to verify its outstanding real-world performance.

 \item [4)]
The ROG-Map is implemented as a ROS package with carefully engineered work and is open-sourced to promote LiDAR-based motion planning.

\end{itemize}

\section{Related Works}
\label{sec:related}
\subsection{Occupancy Grid Map}

The occupancy grid map (OGM) is a promising navigation map type for robots, capable of distinguishing between occupied, free, and unknown environmental areas through ray casting and handling sensor noise and dynamic objects through probabilistic updates.
Existing methods for implementing occupancy maps can be divided into three main streams: octree-based\cite{hornung2013octomap}, hash table-based\cite{niessner2013real}, and uniform grid-based\cite{zhou2020ego}.
We begin by analyzing these methods from the perspective of time complexity. Octree-based methods have a time complexity of $O(\log n)$ for map operations such as insertion, change, and query, where $n$ represents the number of nodes in the tree. In contrast, hash table-based methods have a theoretical time complexity of $O(1)$, but the worst-case time complexity of the operations on the map is $O(n)$ due to hash conflicts \cite{ericson2004real}. Moreover, since both occupied and free grids are maintained on OGMs, the number of hash conflicts increases with a denser map at a higher resolution, which can reduce overall performance. Considering the large number of map operations required in ray casting and obstacle inflation with LiDAR, both of these map types can suffer from working in real-time with onboard computation devices. In contrast, uniform grid-based OGMs \cite{zhou2020ego} ensure that the time complexity of all map operations is $O(1)$ under all circumstances and they are typically several times faster than octree-based and hashing-based methods in map operations.
Next, we focus on the space complexity of those methods. Hashing-based and octree-based occupancy grid maps (OGMs) exhibit a space complexity of $O(n)$, where $n$ denotes the number of nodes in the map. Octree-based methods further optimize memory usage by merging adjacent grids into larger ones, resulting in higher memory efficiency compared to hashing-based methods. Notably, both methods demonstrate a space complexity that remains independent of the map size. In contrast, uniform grid-based methods allocate memory for all grids in the map and have a space complexity of $O(m)$, where $m$ represents the map size. As the memory consumption scales linearly with map size, this method is not suitable for larger-scale missions.
To address these limitations, we propose ROG-Map, a uniform grid-based approach for efficient map operations. To address the space complexity issue, we introduce a zero-copy map sliding strategy that keeps a robocentric local map with a fixed size. In this way, memory consumption of ROG-Map is constant, making it well-suited for missions in large-scale environments.

\subsection{Obstacle Inflation}
In robot motion planning, inflating obstacles is a widely used technique for generating the robot's configuration space \cite{ren2022bubble,liu2018search}. Modeling the robot as a point mass in the configuration space can significantly simplify and accelerate the motion planning algorithm. Traditional obstacle inflation algorithms, such as the one presented in \cite{zhou2020ego}, compute the bounding box of all point clouds in each input frame. After ray casting and probability updating, they traverse all the grids in the bounding box, marking occupied grids and their neighbors as \inflatedoccupied, and the remaining grids as non-\inflatedoccupied. However, for LiDAR point clouds, the bounding box is often large, making map traversal and inflation time-consuming. To address this issue, Li \etal~proposed an incremental update algorithm based on grid state changes in \cite{li2023fiimap}. This algorithm uses a Rising Queue and a Falling Queue to track grids that change from non-\occupied~to \occupied~and from \occupied~to non-\occupied, respectively. The algorithm sets all grids in the Rising Queue and their neighbors as \inflatedoccupied. Then, for each grid in the Falling Queue, it traverses all its neighbors and the neighbors' neighbors to decide if the grid and its neighbor should be set as non-\inflatedoccupied. However, this process has a worst-case complexity of $O(n^2)$, where $n$ is the number of changed grids. In contrast, we propose a novel incremental update scheme that ensures $O(n)$ computation complexity for all cases. Our proposed method reduces the number of traversed grids by \SI{70}{\percent}$\sim$\SI{97}{\percent} on public datasets compared to \cite{li2023fiimap}, significantly accelerating the obstacle inflation process.

\section{Occupancy Grid Map}
\label{sec:occ}
In this section, we introduce the fundamental concept of the occupancy grid map, including the probabilistic update process and the definition of grid states.

At the $k$-th update, the occupancy grid map (OGM) takes the LiDAR position $\mathbf x_k\in \mathbb R^3$ and a scan of LiDAR points $\mathcal P_k$ as input and fuses the measurements using Bayesian updating\cite{hornung2013octomap, mora1985highreso}. If a LiDAR point falls in a grid, it is considered a $\mathtt{hit}$, while if the LiDAR beam passes through a grid, it is considered a $\mathtt{miss}$. Assuming the map update process is Markovian, we estimate the occupancy probability of a grid $\mathbf n$ given the history of measurements up to time $k$, $P(\mathbf n|\mathbf x_{1:k},\mathcal P_{1:k})$, denoted as $P_{1:k}(\mathbf n)$ for brevity, by
\begin{equation}
\label{eq:prob}
\begin{aligned}
P_{1:k}(\mathbf n)
=&
 \left[
 1+\mathbf P
 \right]^{-1}\\
 \mathbf P =& \frac{1-P_{k}(\mathbf n)}{P_k(\mathbf n)}\frac{1-P_{1:k-1}(\mathbf n)}{P_{1:k-1}(\mathbf n)}\frac{P(\mathbf n)}{1-P(\mathbf n)}
\end{aligned}
\end{equation}
where $P(\mathbf n)$ is a prior probability, which is commonly assumed as $P(\mathbf n) = 0.5$ to indicate that the map has no prior information of the occupancy state. By using the log-odd notation 
\begin{equation}
L_{(\cdot)}(\mathbf n)= \log\left(\frac{P_{(\cdot)}(\mathbf n)}{1-P_{(\cdot)}(\mathbf n)}\right), 
\end{equation}
the equation (\ref{eq:prob}) can be rewritten as 
\begin{equation}
\label{eq:log_odds}
L_{1:k}(\mathbf n) = L_{1:k-1}(\mathbf n) + L_k(\mathbf {n})
\end{equation}
Let the log-odds value of a hit be denoted as $l_\mathrm{hit}$ and that of a miss as $l_\mathrm{miss}$, then $L_k(\mathbf n)$ can be computed as follows:

\begin{equation}
 L_k(\mathbf n) = n_\mathrm{hit} \cdot l_\mathrm{hit} + n_\mathrm{miss} \cdot l_\mathrm{miss}
\end{equation}
where $n_\mathrm{hit}$ and $n_\mathrm{miss}$ represent the number of hits and misses, respectively, for the grid at the $k$-th update. {It should be noted that $l_\mathrm{miss}$ is negative when $p_\mathrm{miss} < 0.5$, which means that the occupancy probability decreases when a LiDAR beam passes through a grid.}
\begin{figure}[htbp]
 \centering
 \includegraphics[width=0.43\textwidth]{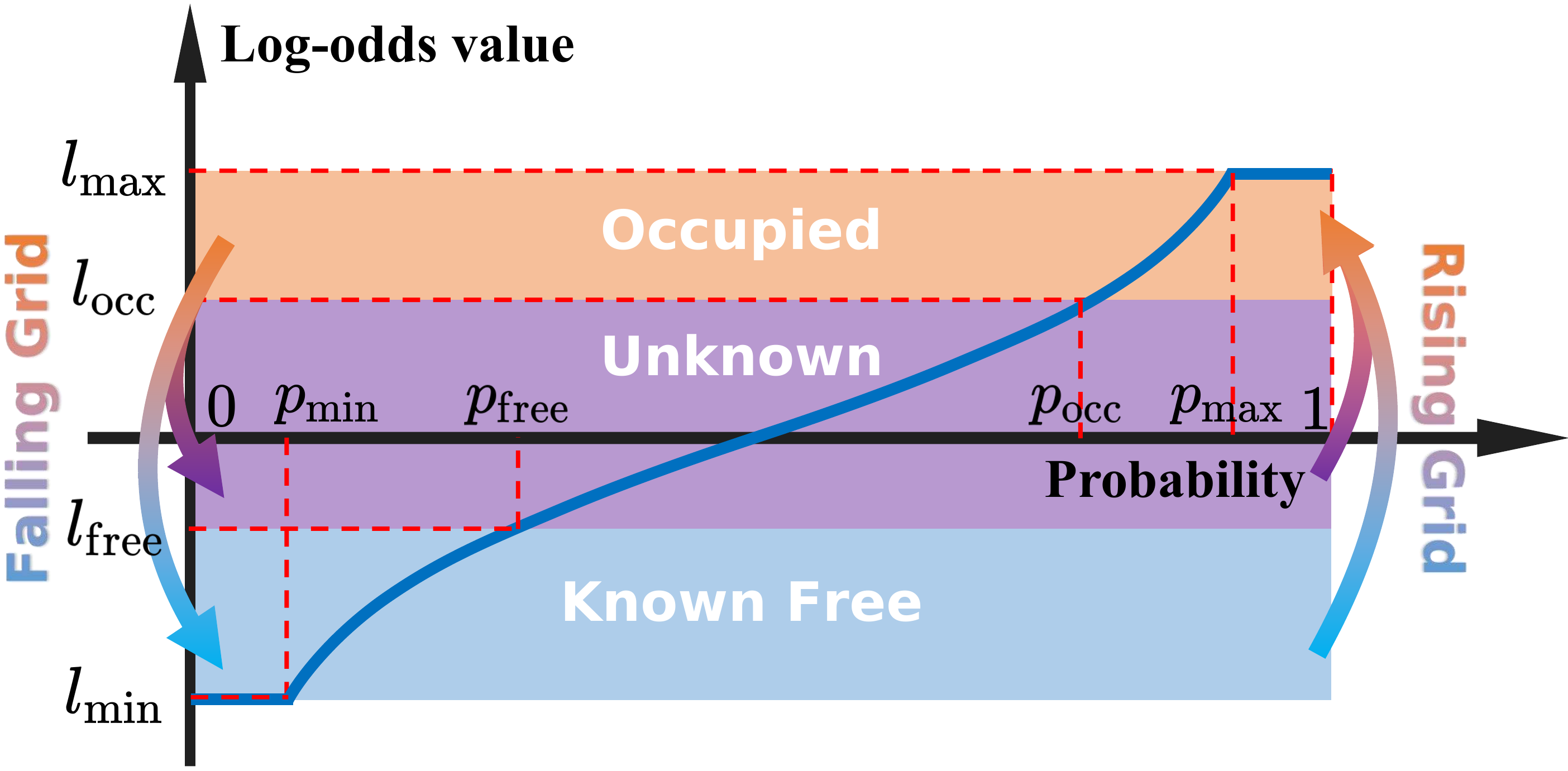}
 \caption{{The log-odds function and the grid state corresponding to the log-odds value. A grid changes from \unknown~or \knownfree~to \occupied~is called a rising grid (RG), while a grid changes from \occupied~to \unknown~or \knownfree~is called a failling grid (FG).}}
\label{fig:logodds}
\end{figure}

Since the log-odds operation is bijective, we only save $L_{1:k}(\mathbf n)$ in our map. As described in \cite{yguel2008update}, to ensure the adaptability in both static and dynamic enviroments, we use a clamping update policy which defines an upper and lower bound on $P_{1:k}(\mathbf n)$. 
In this way, the occupancy state is estimated by
\begin{equation}
\begin{aligned}
&L_t = L_{1:k-1}(\mathbf n) + L_k(\mathbf n)\\
&L_{1:k}(\mathbf n)=\max\left(\min( L_t,l_{\max}),l_{\min}\right)
\end{aligned}
\end{equation} 
{where $l_{\max}$ and $l_{\min}$ are the upper and lower bound on the log-odds value. 
\begin{equation}
 \mathtt{state}(\mathbf n) = 
 \begin{cases}
\mathtt{KnownFree } & l_{\min} \leq {L(\mathbf n)} < l_\mathrm{free}\\
\mathtt{Occupied} & l_\mathrm{occ} \leq {L(\mathbf n)} \leq l_{\max}\\
\mathtt{Unknown} &\mathrm{otherwise}\\
 \end{cases}
\end{equation}

The clamped log-odds function is shown in Fig.~\ref{fig:logodds}. A grid whose state changes from \unknown~or \knownfree~to \occupied~is called a rising grid (RG), and a grid whose state changes from \occupied~to \knownfree~or \unknown~is called a falling grid (FG). The concept of rising and falling grid detection is essential for our novel incremental updating algorithms, which will be detailed in Sec.~\ref{sec:incremental}

\section{The ROG-Map}

The ROG-Map is a uniform grid-based OGM with two main parts in its update process, as shown in Alg.\ref{alg:over}:
\textbf{1) {Map sliding}} (Sec.~\ref{sec:robocentric}), which involves updating the local map origin and resetting the memory outside of the local map, and \textbf{2) Map update} (Sec.~\ref{sec:incremental}), which includes probabilistic update and incremental obstacle inflation.

\begin{algorithm}[h]
\caption{Overview of ROG-Map Update}
\small
\label{alg:over}
\textbf{Notation}: The robot's position $\mathbf{x}_k$; Input point cloud $\mathcal{P}_k$; The update candidate queue $\mathcal C$; Map sliding threshold $d$; Current local map center $\mathbf o$.\\
\KwIn{
$\mathbf{x}_k$, $\mathcal{P}_k$
}
\BlankLine
\SetKwProg{Alg}{Algorithm}{}{}
\Alg{}{
\texttt{/* === Map Sliding === */}\\
\If{$\|\mathbf o - \mathbf x_k\| > d$}{
$\mathtt{UpdateLocalMapOrigin}(\mathbf{x}_k)$\;
$\mathtt{ResetMemoryOutsideMap}(\mathbf{x}_k)$\;
}
\texttt{/* === Map Update === */ }\\
$\mathcal C = \mathtt{Raycasting}(\mathbf{x}_k,\mathcal{P}_k$)\;
$\mathtt{IncrementalInflation}(\mathcal C)$\;
}
\textbf{End Algorithm}
\end{algorithm}

\subsection{Robocentric Local Maps}
\label{sec:robocentric}
In most robotic navigation missions, the robots only need to consider the surrounding environment. Our local data storage is implemented inspired by \cite{Fankhauser2016GridMapLibrary} but further extends it to the three-dimensional case. We leverage a three-dimensional circular buffer for non-destructive shifting of the map's origin (\eg, the robot's position) without copying any data in the memory. 

Assume the local map size is $\mathbf s = (s_x,s_y,s_z)\in \mathbf Z_+^3$ and the discrete resolution is $r$. Without loss of generality, we assume all elements of $\mathbf s$ are odd. All data is saved in an array. For an arbitrary 3D point $\mathbf p = (p_x,p_y,p_z)\in\mathbb R^3$, we define its global index $\mathbf i^g = (i^g_x,i^g_y,i^g_z)$, and its local index $\mathbf i^l = (i^l_x,i^l_y,i^l_z)$. In each dimension $k\in\{x,y,z\}$, the indexes can be computed by:
\begin{equation}
\begin{aligned}
 i^g_k =& \mathtt{round}( p_k / r)\\ 
 i^t_k =& i_k^g~\mathtt{mod}~s\\
 i^l_k =& \mathtt{normalize}(i^t_k)
\end{aligned}
\end{equation}
where $i^t_k$ is an intermediate variable and $\mathtt{normalize}()$ is an normalize function to ensure $i^l_k \in \left[0,s_k-1\right]$:
\begin{equation}
\mathtt{normalize}(x,s) = 
\begin{cases}
x- \lfloor\frac{s}{2}\rfloor & x > \lfloor \frac{s}{2} \rfloor\\
x + \lfloor\frac{s}{2}\rfloor & - \lfloor\frac{s}{2}\rfloor\leq x\leq \lfloor\frac{s}{2}\rfloor\\
x+3\lfloor\frac{s}{2}\rfloor & x < - \lfloor \frac{s}{2}\rfloor\\
\end{cases}
\end{equation}
where $\lfloor\cdot\rfloor$ is the floor function. Then we define the indexing function:
\begin{equation}
\mathtt{toAddress}(\mathbf i^l) =i^l_x \cdot s_y\cdot s_z +i^l_y \cdot s_z+i^l_z
\end{equation}

The indexing process is shown in Fig.~\ref{fig:indexing}. With the above mentioned formulation, we can calculate the address of any point without knowing the local map's origin or the map boundary, making it suitable for robotic motion planning in unbounded scene.

\begin{figure}[htbp]
 \centering
 \includegraphics[width=0.48\textwidth]{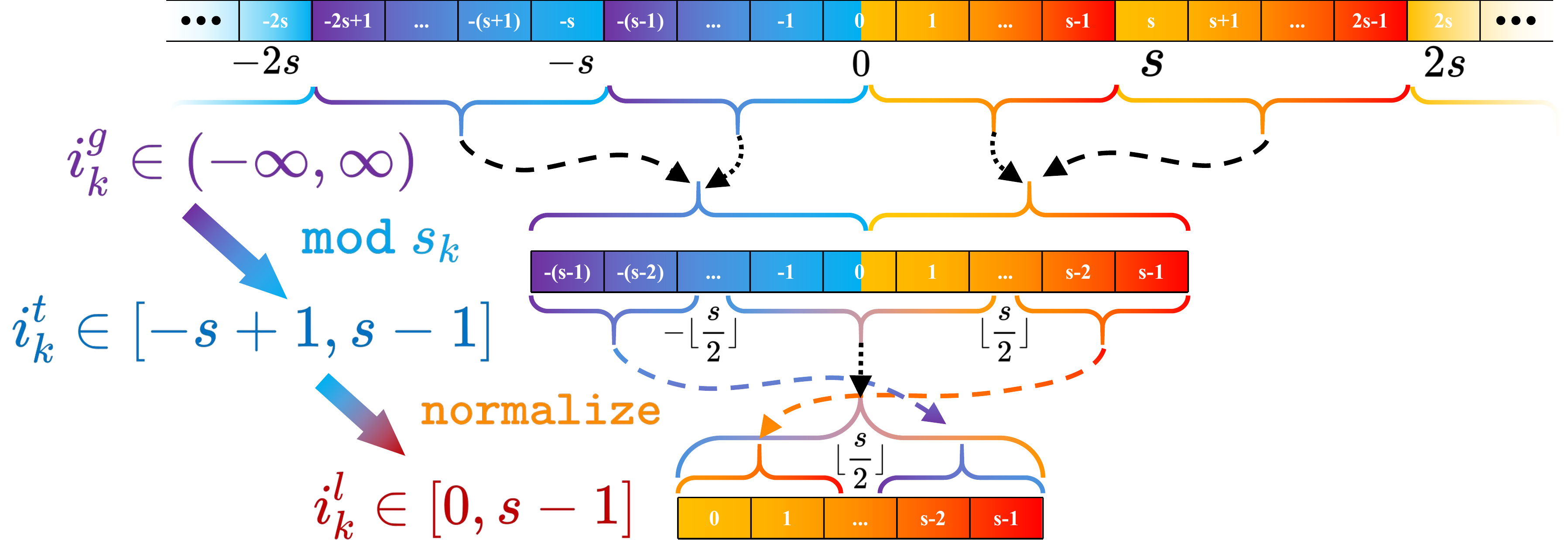}
 \caption{
 The indexing process, which maps the unbounded global index $i_k^g$ to a unique local index $i_k^l$ without knowing the local map origin.
 }
 \label{fig:indexing}
\end{figure}

The relationship between the local map and memory is illustrated in Fig.~\ref{fig:rolling}, we only show the map at the robot's height to ease the visualization. As the origin of the map (i.e., the robot) moves, the area within the yellow dashed box at $t_{k-1}$ goes out of the local map at $t_k$, hence its corresponding space in the memory is reset. The reset space is then used to save newly encountered area shown in the green dashed box. The red dashed boxes indicate that the addresses of the cells remain unchanged in memory before and after the local map's movement. To avoid frequent memory reseting, the map sliding operation for updating the map center and resetting the memory outside the map is only executed when the distance between the robot and the current center of the local map is greater than a threshold $d$. It is worth noticing that the mapping from the global to the local index is a surjective map, and each local index corresponds to multiple global indexes. However, the global index can be uniquely determined from the local index given the map's origin and the range of the local map. 

\begin{figure}[htbp]
 \centering
 \includegraphics[width=0.43\textwidth]{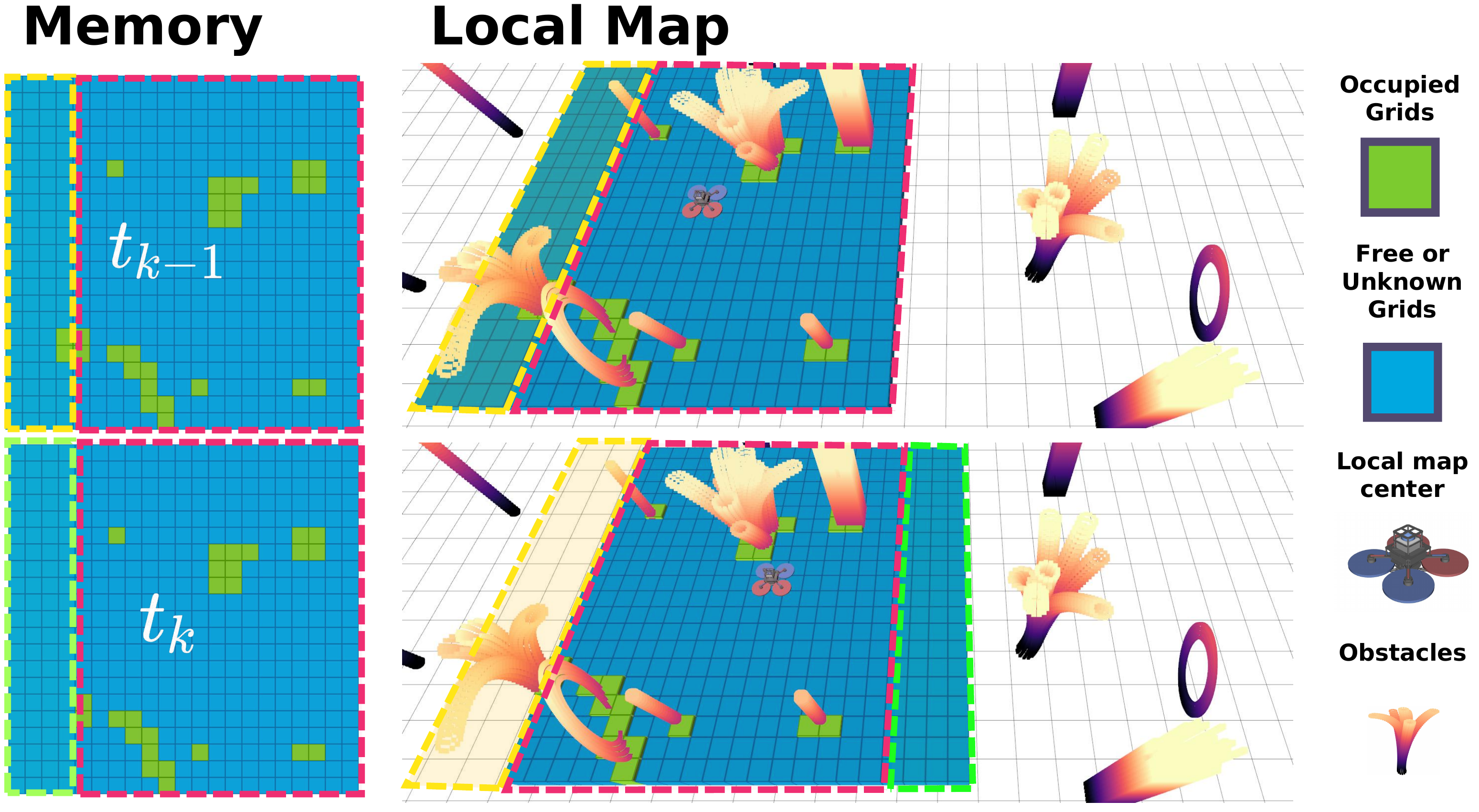}
 \caption{The visualization of the local map and memory at the robot's height. The \unknown~and \knownfree~grids are colored in blue, while the \occupied~grids are colored in green. 
The dashed boxes in the same colors indicate the same area in memory and local map.
}
\label{fig:rolling}
\end{figure}

\subsection{Probabilistic Update and Incremental Inflation}
\label{sec:incremental}

The map updating process is presented in Alg. \ref{alg:update}. For the $k$-th scan of point cloud $\mathcal{P}_k$ and the associated position of the LiDAR $\mathbf x_k$, the first step in the map updating process is ray casting, as shown in Line \ref{alg:update:ray casting}. To accomplish this, we use a fast voxel traversal algorithm \cite{amanatides1987fast} to traverse all grids between each LiDAR point and $\mathbf x_k$. Since multiple rays may cross the same grid, we use a cache $\mathcal C$ to save all traversed grids and the number of hits $c_\mathrm{hit}$ and misses $c_\mathrm{miss}$ following \cite{hornung2013octomap}. This batch update approach significantly reduces the number of map operations in the following probabilistic update process. The ray casting process is encapsulated in the function $\mathcal C = \mathtt{ray casting}(\mathbf x_k,\mathcal P_k)$. After performing ray casting, we process all candidate grids in $\mathcal{C}$ (Lines~\ref{alg:update:cadidate_start}-\ref{alg:update:cadidate_end}). The log-odds probability is updated using (\ref{eq:log_odds}) in Line~\ref{alg:update:update_log_odds}.

\begin{algorithm}[h]
\caption{Local Map Update and Obstacle Inflation}
\small
\label{alg:update}
\textbf{Notation}: The robot's position $\mathbf{x}_k$; Input point cloud $\mathcal{P}_k$; The updating candidates queue $\mathcal{C}$; The spherical update list $\mathcal L$.\\
\BlankLine
\SetKwProg{Alg}{Algorithm}{}{}
\Alg{}{
$\mathcal{C} \leftarrow \mathtt{ray casting}(\mathbf{x}_k, \mathcal{P}_k)$\;
\label{alg:update:ray casting}
\ForEach{$\mathbf{n} \in \mathcal{C}$}{\label{alg:update:cadidate_start}

$\mathbf{n}_{\mathrm{last}} \leftarrow \mathbf{n}$\;

$\mathbf n \leftarrow \mathtt{UpdateProbability}(\mathbf n)$\label{alg:update:update_log_odds}\;



\If{$\mathtt{IsRaisingGrid}(\mathbf{n}_{\mathrm{last}}, \mathbf{n})$}{
$\mathtt{UpdateNeighborCounter}(\mathbf{n}, +1)$\;
}

\If{$\mathtt{IsFallingGrid}(\mathbf{n}_{\mathrm{last}}, \mathbf{n})$}{
$\mathtt{UpdateNeighborCounter}(\mathbf{n}, -1)$\;
}
}
}\label{alg:update:cadidate_end}
\textbf{Algorithm End}\\

\BlankLine

\SetKwProg{Fn}{Function}{}{}
\Fn{$\mathtt{UpdateNeighborCounter(\mathbf n, c)}$}{
\ForEach(){$\mathbf p \in \mathcal L$}{
 $\mathbf n_{\mathrm{temp}} = \mathbf n + \mathbf p$\;\label{alg:udpate:nei}
 $\mathbf n_{\mathrm{temp}}$.inflationCounter += $c$\;
 }
}
\textbf{End Function}\\
\end{algorithm}

\begin{table*}
\begin{threeparttable}[ht]
\vspace{-0.3cm}
\scriptsize
\centering
\caption{Benchmark Comparison}
\setlength{\tabcolsep}{1mm}
\label{tab:benchmark}
\begin{tabular}{@{}lcccccccccccc@{}}
\midrule
& \multicolumn{6}{c}{New College (Ave. 156.2 points per frame)} & \multicolumn{6}{c}{HKU RSC (Ave. 23157.6 points per frame)} \\ \cmidrule(r){2-7} \cmidrule(r){8-13} 
&$t_\mathrm{tot}$ (ms)&$t_\mathrm{u}$ (ms) & $n_\mathrm{inf}$ & $t_\mathrm{inf}$ (ms) & $t_\mathrm{q}$(ms) & $m$ (MB) &
$t_\mathrm{tot}$ (ms) &$t_\mathrm{u}$ (ms) & $n_\mathrm{inf}$ & $t_\mathrm{inf}$ (ms) & $t_\mathrm{q}$(ms) & $m$ (MB) \\ \midrule

OctoMap\cite{hornung2013octomap} & 2.563 & 0.983 & 34991.262 & 1.625 & 61.463 & 640.809 
& 2255.409& 1502.598 & 27258042.859 & 752.811 & 182.306 & \textbf{186.7}\\
HashMap 
& 1.719& 0.218& 34991.262 & 1.501 & 48.771 & 2101.527 
& 2010.611& 1165.105 & 27258042.859 & 845.506 & 152.322 & 2219.953\\
UniformMap\cite{zhou2020ego} 
& 0.224 & \textbf{0.095} & 34991.262 & 0.129 & \textbf{21.688} & 3519.195 
& 133.514& 70.897& 27258042.859 & 62.617 & 23.896 & 11643.707\\
FIIMap\cite{li2023fiimap} 
& 0.119& 0.106 & 1617.898 & 0.013 & 22.708 & 3519.230 
& 267.990&\textbf{63.759} & 1346163.411 & 204.231\tnote{1} & 22.104 &11678.153\\
ROG-Map 
& \textbf{0.101}&0.097 & \textbf{305.917} & \textbf{0.004} & 22.981 & \textbf{64.387} 
& \textbf{68.187}& 67.416& \textbf{36059.518} & \textbf{0.771} & \textbf{20.467} & 1046.243 \\ \bottomrule 
\end{tabular}
 \begin{tablenotes}
 \item [1] {Following \cite{li2023fiimap}, we implemented incremental inflation for FIIMap using a queue structure. Although the number of accessed grids is smaller, using a queue instead of an array significantly reduces the cache hit rate, resulting in a longer overall computation time.} 
 \end{tablenotes}
\end{threeparttable}%
\end{table*}

Then we perform obstacle inflation by detecting the rising grids (RGs) and falling grids (FGs) defined in Sec. \ref{sec:occ}. For each grid in the local map, an integer is used to count how many neighbors of this grid are \occupied, with the counter initially set to zero. In order to perform obstacle inflation, it is necessary to traverse all neighbors within the inflation distance of the current grid.
Since the relative positions of the neighbors and the current grid are known, we can build a look-up table $\mathcal{L}$ to save all neighbor's relative indexes and obtain the neighbor's index using Line~\ref{alg:udpate:nei}. For each RG, all neighbor counters should increase by 1, and for each FG, the counter should decrease by 1. This way, when a grid's inflation count is greater than or equal to one, it indicates that at least one of its neighbors is \occupied~and therefore, the grid is considered as \inflatedoccupied, as shown in Fig.~\ref{fig:inf}.Our incremental inflation strategy has a time complexity of $O(n)$ in all circumstances, where $n$ is the number of changed grids (both RGs and FGs). Compared to \cite{li2023fiimap}, which has a worst-case complexity of $O(n^2)$, the proposed method reduces the number of grid traversals in obstacle inflation by up to \SI{70}{\percent} in the New College Dataset and \SI{97}{\percent} in HKU RSC (detailed in Sec.~\ref{sec:bench}).

\begin{figure}[htbp]
 \centering
 \includegraphics[width=0.43\textwidth]{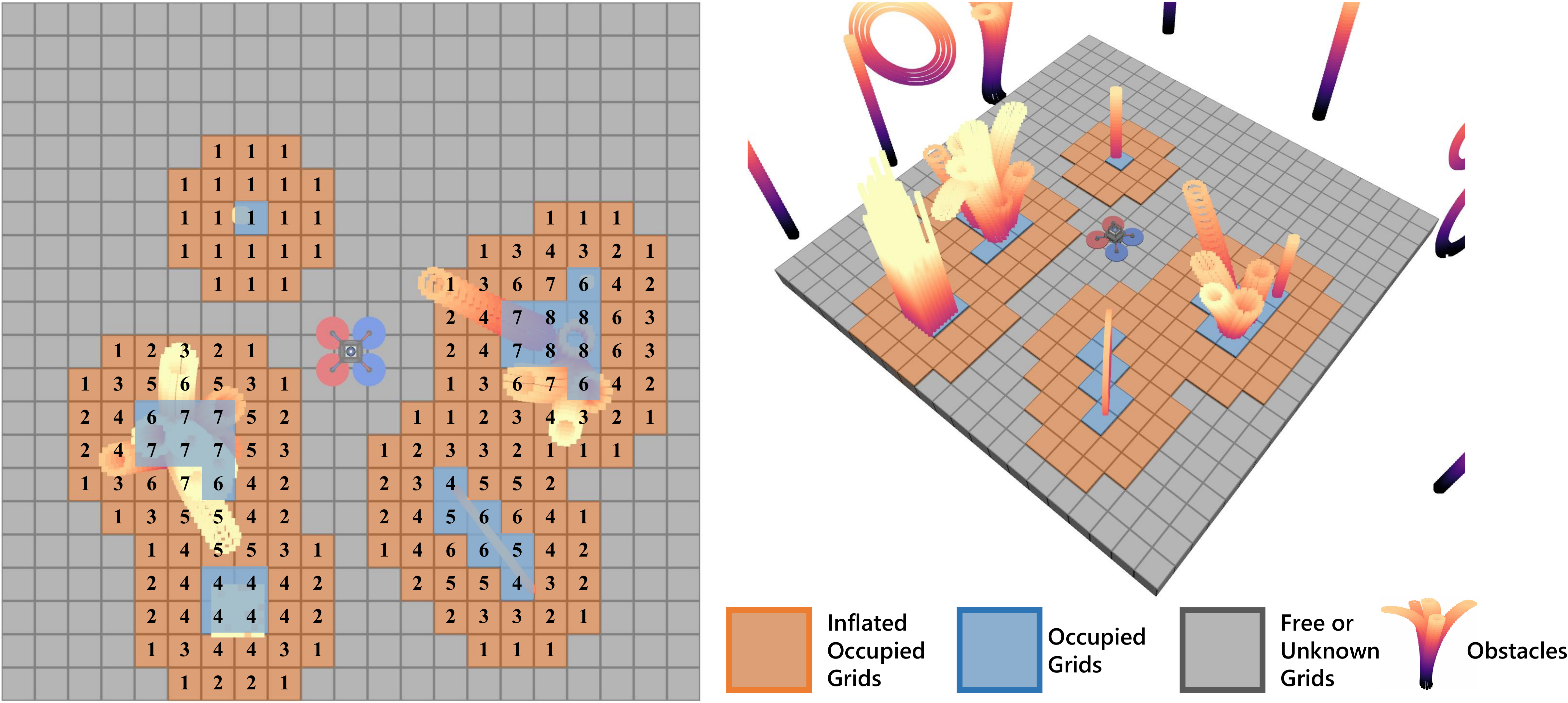}
 \caption{The visualization of $\mathtt{inflationCounter}$ in the local map at the robot's height. The orange and blue grids are \inflatedoccupied~and the gray grids are \knownfree~or \unknown.
}
\label{fig:inf}
\end{figure}

 \section{Benchmark Comparison}
 \label{sec:bench}
 
In this section, we compare the proposed ROG-Map with four different types of baselines: \textbf{1) OctoMap} \cite{hornung2013octomap}, which is an octree-based OGM; \textbf{2) HashMap}, which is a hash table-based OGM. Since we could not find a suitable open-source hash-table based occupancy grid map, we implemented one based on \cite{niessner2013real, r3live_pp, lin2023immesh}; \textbf{3) UniformMap} \cite{zhou2020ego}, which is an OGM with uniform grids and fixed map origin; and \textbf{4) FIIMap} \cite{li2023fiimap}, {a uniform grid-based OGM that features an incremental obstacle inflation approach. As the source code for \cite{li2023fiimap} is not publicly available, we implemented this approach by integrating the incremental inflation algorithm described in the paper into the UniformMap.} We compare the average computation time per frame $t_\mathrm{tot}$ (including map update, and obstacle inflation), the computation time of probabilistic update $t_u$, the number of operated grids in obstacle inflation $n_\mathrm{inf}$, the time consumed for 100,000 random queries $t_\mathrm{q}$, and the memory consumption $m$. It should be noted that \textbf{1) - 3)} inflate obstacles by traversing a local updating box and inflating all occupied grids within it, following \cite{zhou2020ego}.

We compare the aforementioned methods on two datasets: \textbf{1) New College Dataset} an outdoor dataset provided by \cite{hornung2013octomap}\footnote{http://ais.informatik.uni-freiburg.de/projects/datasets/octomap/}. The scene size is \SI{250}{\meter}$\times$\SI{161}{\meter}$\times$\SI{33}{\meter}. The resolution of all methods is set to \SI{0.2}{m} following \cite{hornung2013octomap}. The obstacle inflation distance is set to \SI{0.2}{m}, and the maximum raycast distance is set to \SI{20}{m}. \textbf{2) HKU RSC}\footnote{https://github.com/hku-mars/MARSIM}: an indoor dataset proposed by \cite{kong2022marsim}. The size of the dataset is \SI{74}{m}$\times$ \SI{42}{m}$\times$\SI{20}{m}, and the mapping resolution is set to \SI{0.05}{m} with a maximum raycast distance of \SI{20}{m}. The obstacle inflation distance is set to \SI{0.2}{m}. 

Since ROG-Map's performance is independent of the map size, it only needs to be set based on the available memory size. In both of the two tests, we set it considering the distance used for ray casting (\SI{20}{m}), which is \SI{40}{m}$\times$\SI{40}{m}$\times$\SI{12}{m}.

The benchmark results are presented in Table~\ref{tab:benchmark}. 
ROG-Map performs similarly to other uniform grid-based approaches in probabilistic update and random query, but is several times faster than hashing-based and octree-based methods. In terms of obstacle inflation, ROG-Map requires only \SI{0.13}{\percent}$\sim$\SI{0.87}{\percent} of the map operations of traversal-based methods (\ie, OctoMap, HashMap, and UniformMap) and only \SI{2.67}{\percent}$\sim$\SI{18.12}{\percent} compared to FIIMap, resulting in significantly less time for obstacle inflation and overall map updates. Notably, the HKU RSC dataset was captured at a rate of \SI{10}{Hz} (\ie, \SI{100}{ms} per frame), and ROG-Map only requires an average of \SI{68.187}{ms} to process, highlighting its real-time capability at a high resolution of \SI{0.05}{m}, while none of the other approaches can run in real-time.
From the perspective of memory consumption, OctoMap is the most memory-efficient, followed by HashMap. Uniform grid-based methods have the same space complexity, which increases linearly with the map size and is thus inefficient compared to OctoMap and HashMap. ROG-Map maintains a robocentric local map, resulting in a constant space complexity, making it suitable for large-scale missions with acceptable memory consumption.

\section{Applications}

The ROG-Map was integrated into a LiDAR-based quadrotor platform to assess its real-world performance. The on-board computing unit was an Intel NUC equipped with a CPU i7-10710U. The platform weighed \SI{1.5}{\kilogram} and boasted a thrust-to-weight ratio over $4.0$.

Our perception module utilized the Livox Mid360 LiDAR and Pixhawk's built-in IMU, running a modified version \cite{zhu2022decentralized} of FAST-LIO2\cite{xu2022fast}, which provided high-accuracy state estimation at a frequency of \SI{100}{\hertz} and point clouds at a frequency of \SI{50}{\hertz}. We utilized the method proposed in \cite{zhu2022robust} to calibrate the extrinsic and time-offset between the LiDAR and IMU. For trajectory tracking control, we employed an on-manifold model predictive controller proposed in \cite{lu2022manifold}. To achieve collision avoidance, we utilized a modified version of our previous work \cite{ren2023online} as the local planner. Specifically, we employed ROG-Map to identify collision-free paths and generate safe flight corridors.

\begin{figure}[htbp]
 \centering
 \includegraphics[width=0.47\textwidth]{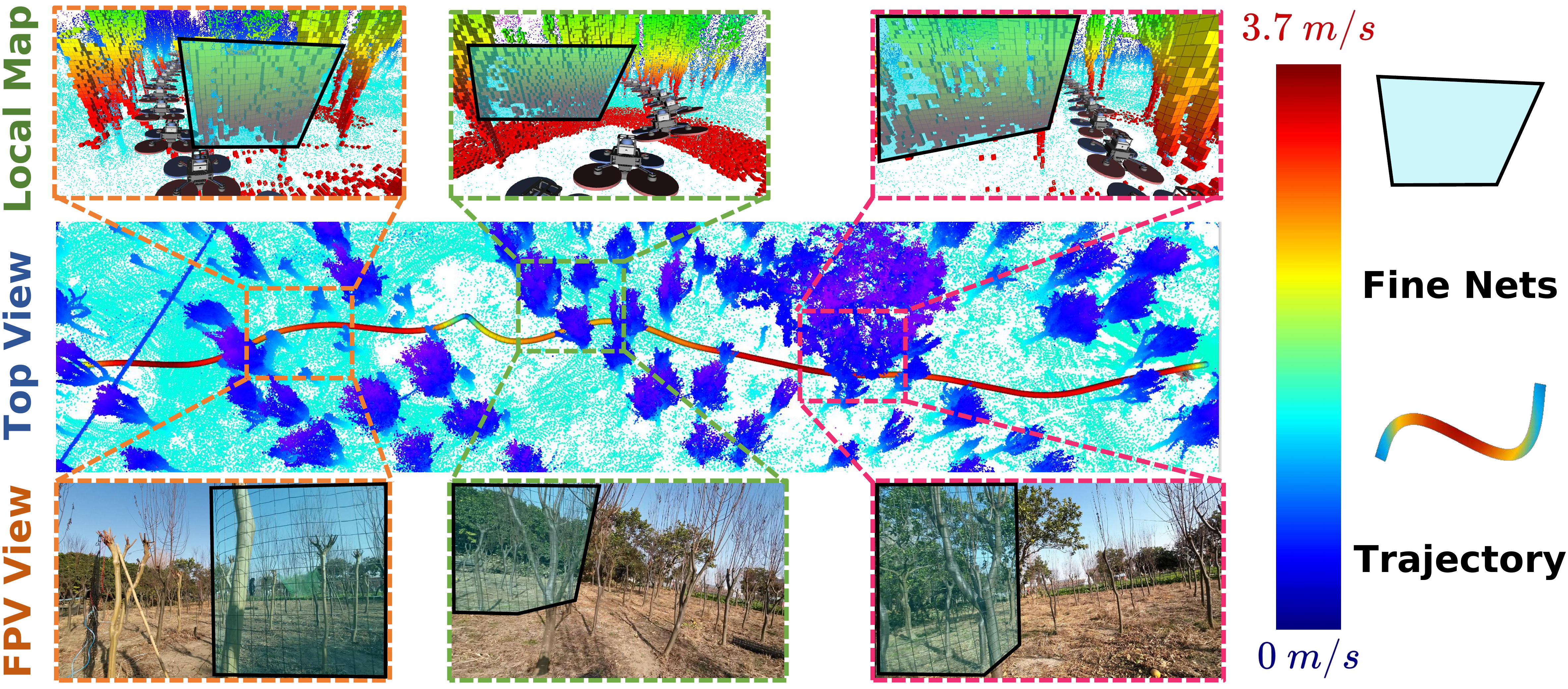}
 \caption{The \textit{Test1} was conducted in a small dense forest of short trees where three fine nets made of \SI{3}{mm} thin metal wires was placed on the flight course. Using ROG-Map, the autonomous quadrotor successfully fly through the scene at a maximum speed over \SI{3.7}{\mps}.}
\label{fig:test1}
\end{figure}

We conducted a total of eight successful experiments, which were divided into two different categories. Due to space constraints, we only present two of these tests and other tests can be found in the attached video\footnote{https://youtu.be/eDkwGXCea7w}. In all of the tests, ROG-Map was set to a size of \SI{30}{m}$\times$\SI{30}{m}$\times$\SI{6}{m} with a resolution of \SI{0.05}{m}. The local map update range was set to \SI{15}{m}.

The first experiment, denoted as \textit{Test 1}, was conducted in a dense forest, as depicted in Fig.~\ref{fig:test1}. We placed thin metal nets to create a more challenging environment. Using ROG-Map, the drone successfully detected the nets and avoided collisions.

In \textit{Test 2}, we presented a multi-waypoints inspection mission in a large-scale scene. The inspection goal was specified by the user, and the drone automatically navigated and avoided collisions while moving toward the designated goal positions. The total travel distance was \SI{502.23}{m}. The drone successfully traversed all waypoints without a crash, demonstrating the usefulness of our local map-shifting strategy. The generated point cloud and the path of the drone are shown in Fig.~\ref{fig:test2}.

\begin{figure}[htbp]
 \centering
 \includegraphics[width=0.45\textwidth]{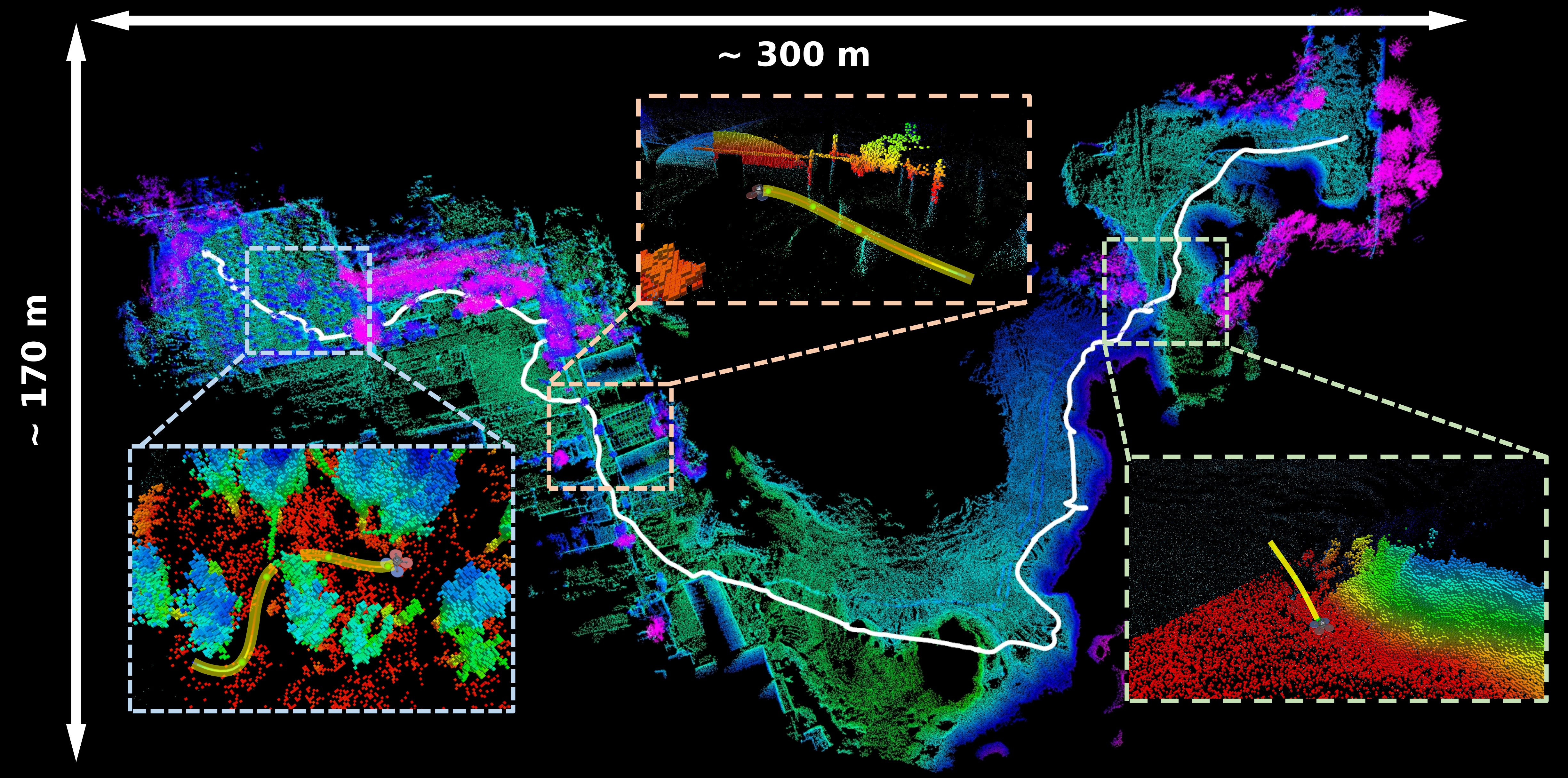}
 \caption{The point cloud view of \textit{Test2}. The planned trajectory is highlighted in yellow. The total distance covered during the autonomous flight was \SI{502.23}{m}. 
}
\label{fig:test2}
\end{figure}

\begin{figure}[htbp]
\centering 
\includegraphics[width=0.48\textwidth]{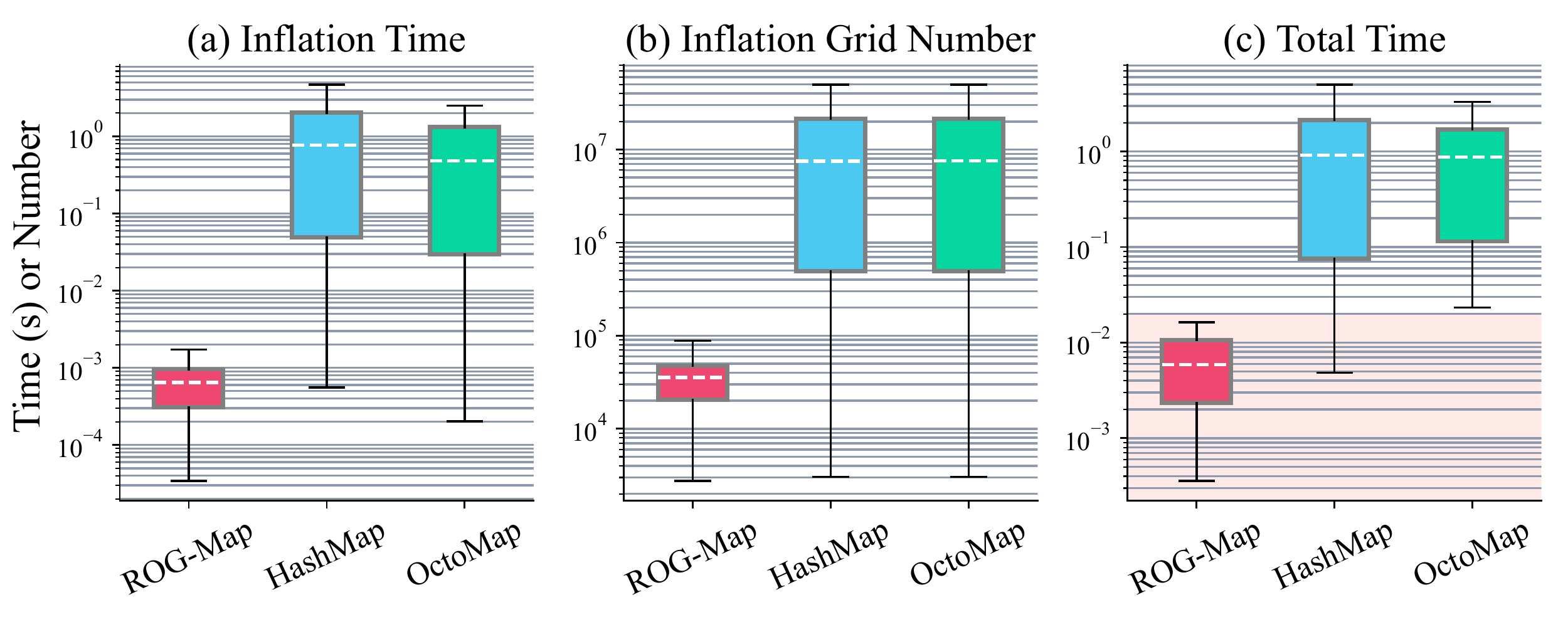}
 \caption{The computation time of ROG-Map in real-word flight and the baseline running the recorded data on the same computer. The red shaded area in (c) indicated the total time is less than \SI{20}{ms}, showing ROG-Map can work with LiDAR input in real-time.
}
\label{fig:real}
\end{figure}

The computation time of ROG-Map is presented in Fig~\ref{fig:real}. We also run the flight dataset of \textit{Test 2} using Octomap and Hashmap. We didn't test UniformMap since the memory consumption is beyond the system memory on the onboard computer. As the point cloud input has a frequency of \SI{50}{Hz} (i.e., \SI{20}{\ms} per frame), neither HashMap nor Octomap can achieve real-time as shown if Fig~\ref{fig:real}(c), where the red shaded area is computation time less than \SI{20}{ms}. Our \textbf{ROG-Map} only takes an average of \SI{5.96}{ms} in \textit{Test1} and \SI{3.46}{ms} in \textit{Test2} per frame for local map updates, including only \SI{0.172}{ms} in \textit{Test1} and \SI{0.06}{ms} in \textit{Test2} for obstacle inflation, demonstrating its high efficiency for robotic planning missions.

\section{Conclusion and Future Work}

In this paper, we {proposed} ROG-Map as a solution for using occupancy grid maps (OGM) with LiDAR sensors. We {used} a uniform grid-based OGM to maintain a local map surrounding the robot with a zero-copy map sliding strategy to ensure computational and memory efficiency. Furthermore, we {proposed} a novel incremental obstacle inflation method that significantly {reduced} the computation time for inflation and {improved} the overall mapping performance. Benchmark comparisons on various public datasets {demonstrated} its advantages over state-of-the-art (SOTA) methods in terms of computation time and memory consumption. Finally, we {integrated} ROG-Map into a complete quadrotor system and {demonstrated} its capability for large-scene high-resolution LiDAR-based motion planning missions.

One limitation of ROG-Map is that as the robot moves away from a region, the occupancy information about the region is cleared. This spatial forgetting mechanism does not affect the robot's performance in obstacle avoidance. However, for other applications, such as autonomous exploration that requires maintaining information about the entire environment, ROG-Map is no longer suitable. One possible solution is to use sparse data structures to record the global map information while using ROG-Map to achieve real-time updates and maintain local map information. In the future, we plan to explore a hybrid map framework that combines global and local maps to support a broader range of tasks.

\section*{Acknowledgment}
The authors gratefully acknowledge the funding provided by DJI and the equipment support provided by Livox Technology during this project. The authors would also like to express their gratitude to Sifan Tang for recording the experiments and to Longji Yin, Yuhan Xie and Fanze Kong for their valuable discussions.

{\small
\bibliographystyle{unsrt}
\bibliography{reference}
}

\end{document}